\pdfoutput=1

\documentclass[11pt]{article}

\usepackage[]{acl}
\usepackage{graphicx} 
\usepackage{times}
\usepackage{latexsym}
\usepackage{amsmath, amssymb} 

\usepackage{adjustbox}
\usepackage{xcolor}

\usepackage{multirow}
\definecolor{darkgreen}{rgb}{0.0, 0.5, 0.0}
\definecolor{darkred}{rgb}{0.5, 0.0, 0.0}
\usepackage[T1]{fontenc}

\usepackage[utf8]{inputenc}

\usepackage{microtype}
\usepackage[switch]{lineno}

\title{3MVRD: Multimodal Multi-task Multi-teacher Visually-Rich Form Document Understanding}

\author{
Yihao Ding\textsuperscript{1,2}, 
Lorenzo Vaiani \textsuperscript{3},
Soyeon Caren Han \textsuperscript{1,2 \thanks{
~~Corresponding Author (caren.han@unimelb.edu.au)
}} , 
Jean Lee\textsuperscript{1}, \\
\textbf{Paolo Garza}\textsuperscript{3},
\textbf{Josiah Poon}\textsuperscript{1},
\textbf{Luca Cagliero}\textsuperscript{3}
\\
\textsuperscript{1}The University of Sydney, \textsuperscript{2} The University of Melbourne, \textsuperscript{3} Politecnico di Torino\\
\tt\small 
\{yihao.ding,caren.han,jean.lee,josiah.poon\}@sydney.edu.au, \\ 
\tt\small 
caren.han@unimelb.edu.au, \{lorenzo.vaiani,paolo.garza,luca.cagliero\}@polito.it\\ 
\tt\small 
}

\begin{document}
\maketitle
\begin{abstract}
This paper presents a groundbreaking multimodal, multi-task, multi-teacher joint-grained knowledge distillation model for visually-rich form document understanding. The model is designed to leverage insights from both fine-grained and coarse-grained levels by facilitating a nuanced correlation between token and entity representations, addressing the complexities inherent in form documents. Additionally, we introduce new intra-grained and cross-grained loss functions to further refine diverse multi-teacher knowledge distillation transfer process, presenting distribution gaps and a harmonised understanding of form documents. Through a comprehensive evaluation across publicly available form document understanding datasets, our proposed model consistently outperforms existing baselines, showcasing its efficacy in handling the intricate structures and content of visually complex form documents\footnote{Code: \url{https://github.com/adlnlp/3mvrd}}.
\end{abstract}

\section{Introduction}
Understanding and extracting structural information from Visually-Rich Documents (VRDs), such as academic papers~\cite{publaynet, pdfvqa}, receipts~\cite{cord}, and forms~\cite{funsd,formnlu}, holds immense value for Natural Language Processing (NLP) tasks, particularly in information extraction and retrieval. While significant progress has been made in solving various VRD benchmark challenges, including layout analysis and table structure recognition, the task of form document understanding remains notably challenging. 
This complexity of the form document understanding arises from two main factors: 1) the involvement of two distinct authors in a form and 2) the integration of diverse visual cues. Firstly, forms mainly involve two primary authors: form designers and users. Form designers create a structured form to collect necessary information as a user interface. Unfortunately, the form layouts, designed to collect varied information, often lead to complex logical relationships, causing confusion for form users and heightening the challenges in form document understanding. Secondly, diverse authors in forms may encounter a combination of different document natures, such as digital, printed, or handwritten forms. Users may submit forms in various formats, introducing noise such as low resolution, uneven scanning, and unclear handwriting. Traditional document understanding models do not account for the diverse carriers of document versions and their associated noises, exacerbating challenges in understanding form structures and their components. Most VRD understanding models inherently hold implicit multimodal document structure analysis (Vision and Text understanding) knowledge either at fine-grained~\cite{layoutlmv3, lilt} or coarse-grained~\cite{lxmert, visualbert} levels. The fine-grained only models mainly focus on learning detailed logical layout arrangement, which cannot handle complex relationships of multimodal components, while the coarse-grained models tend to omit significant words or phrases. 
Hence, we introduce a novel joint-grained document understanding approach with multimodal multi-teacher knowledge distillation. It leverages knowledge from various task-based teachers throughout the training process, intending to create more inclusive and representative multi- and joint-grained document representations. 

Our contributions are summarised as follows: 1) We present a groundbreaking multimodal, multi-task, multi-teacher joint-grained knowledge distillation model designed explicitly to understand visually-rich form documents. 2) Our model outperforms publicly available form document datasets. 3) This research marks the first in adopting multi-task knowledge distillation, focusing on incorporating multimodal form document components. 

\begin{table*}[ht]
\small
\centering
\begin{tabular}{cccccc}
\hline
\textbf{Model} & \textbf{Modalities} & \textbf{\begin{tabular}[c]{@{}c@{}}Pre-training\\ Datasets\end{tabular}} & \textbf{\begin{tabular}[c]{@{}c@{}}Pre-training\\ Tasks\end{tabular}} & \textbf{\begin{tabular}[c]{@{}c@{}}Downstream\\ Tasks\end{tabular}} & \textbf{Granularity} \\ \hline
Donut~\citeyearpar{donut} & V & IIT-CDIP & NTP &    DC, VQA, KIE & Token \\
Pix2struct~\citeyearpar{pix2struct} & V & C4 corpus & NTP &     VQA & Token \\
LiLT~\citeyearpar{lilt} & T, S & IIT-CDIP & MVLM, KPL, CAI &      DC, KIE & Token \\
BROS~\citeyearpar{bros} & T, S & IIT-CDIP & MLM, A-MLM &      KIE & Token \\
LayoutLMv3~\citeyearpar{layoutlmv3} & T, S, V & IIT-CDIP & MLM, MIM, WPA &      DC, VQA, KIE & Token \\
DocFormerv2~\citeyearpar{docformerv2} & T, S, V & IDL & TTL, TTG, MLM &       DC, VQA, KIE & Token \\
Fast-StrucText~\citeyearpar{fast-structext} & T, S, V & IIT-CDIP & MVLM, GTR, SOP, TIA &      KIE & Token \\
FormNetV2~\citeyearpar{formnetv2} & T, S, V & IIT-CDIP & MLM, GCL &          KIE & Token \\ \hline
\begin{tabular}[c]{@{}c@{}}3MVRD (Ours)\end{tabular} & T, S, V & \begin{tabular}[c]{@{}c@{}}FUNSD,\\ FormNLU\end{tabular} & \begin{tabular}[c]{@{}c@{}}Multi-teacher\\ Knowledge \\ Distillation\end{tabular} &          KIE & \begin{tabular}[c]{@{}c@{}}Token,\\ Entity\end{tabular} \\ \hline
\end{tabular}
\caption{Comparison with state-of-the-art models for receipt and form understanding. In the \textit{Modalities} column, \textit{T} represents Textual information, \textit{V} represents Visual information, and \textit{S} represents Spatial information.}
\label{tab:models}
\end{table*}

\section{Related Works}
Visually Rich Document (VRD) understanding entails comprehending the structure and content of documents by capturing the underlying relations between textual and visual modalities. 
Several downstream tasks, such as Layout Analysing \cite{docgcn}, Key Information Extraction (KIE) \cite{vies}, Document Classification (DC) \cite{layoutlm}, and Visual Question Answering (VQA) \cite{vdoc}, have contributed to raising the attention of the multimodal learning community as shown by Table~\ref{tab:models}.
In this work, we cope with form documents, whose structure and content are particularly challenging to understand~\cite{SrivastavaMDM20}. Form documents possess intricate structures involving collaboration between form designers, who craft clear structures for data collection, and form users, who interact with the forms based on their comprehension, with varying clarity and ease of understanding. 

\noindent \textbf{Vision-only approaches:} They exclusively rely on the visual representation (denoted by \textit{V} modality in Table~\ref{tab:models}) of the document components thus circumventing the limitations of state-of-the-art text recognition tools (e.g., Donut~\cite{donut} and Pix2struct~\cite{pix2struct}). 
Their document representations are commonly pre-trained using a Next Token Prediction (NTP) strategy, offering
alternative solutions to traditional techniques based on Natural Language Processing. 

\noindent \textbf{Multimodal approaches:} They leverage both the recognised text and the spatial relations (denoted by \textit{T} and \textit{S}) between document components (e.g., LiLT~\cite{lilt} and BROS~\cite{bros}). The main goal is to complement raw content understanding with layout information. 
Expanding upon this multimodal framework, models such as LayoutLMv3~\cite{layoutlmv3}, DocFormerv2~\cite{docformerv2}, Fast-StrucText~\cite{fast-structext}, and, FormNetV2~\cite{formnetv2} integrate the visual modality with text and layout information. These approaches 
are capable of capturing nuances in the document content hidden in prior works. 
To leverage multimodal relations, these models are typically pre-trained in a multi-task fashion, exploiting a curated set of token- or word-based pre-training tasks, such as masking or alignment.

\noindent \textbf{Our approach} aligns with the multimodal model paradigm, distinguishing itself by eschewing generic pre-training tasks reliant on masking, alignment, or NTP. Instead, it leverages the direct extraction of knowledge from \textit{multiple teachers}, each trained on downstream datasets, encompassing \textit{both entity and token levels} of analysis with the proposed intra-grained and cross-grained losses. This enriches the depth of understanding in visual documents, capturing intricate relationships and semantic structures beyond individual tokens.

\begin{figure*}[th]
  \centering
  \includegraphics[width=\linewidth]{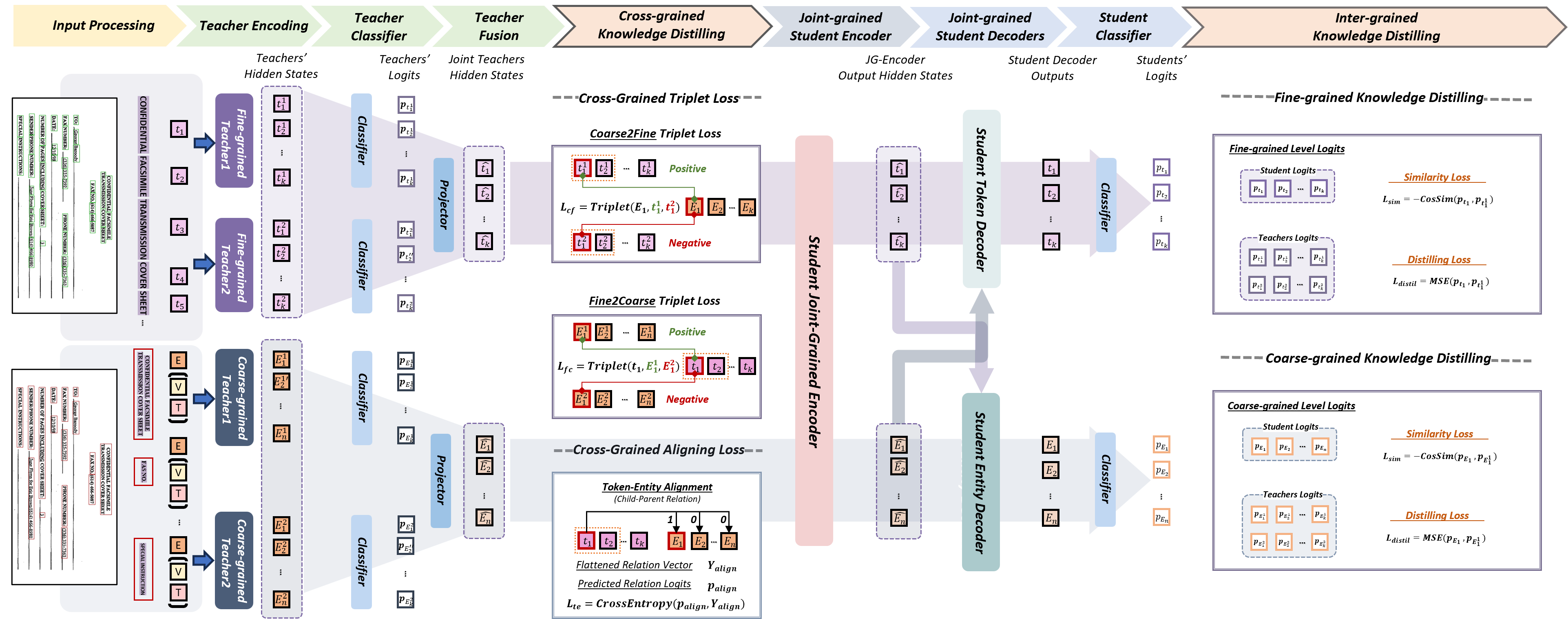}
  \caption{Multimodal Multi-task Multi-teacher Visually-Rich Form Document Understanding (3MVRD). Each section is aligned with the specific colours, Green: Section~\ref{sec:mmmt}, Blue: Section~\ref{sec:jointl}, Orange: Section~\ref{sec:mmm_loss}}
  \label{fig:task_definition}
\end{figure*}

\section{Methodology}
As previously noted, our paper focuses on interpreting visually rich documents, particularly form documents created and used collaboratively by multiple parties. To accomplish this objective, we introduce and employ two tiers of multimodal information: fine-grained and coarse-grained levels, which play a crucial role in understanding the structure and content of an input form page. Note that existing pre-trained visual-language models, whether designed for generic documents, possess implicit knowledge on either fine-grained or coarse-grained aspects. Hence, we propose an approach that harnesses knowledge from diverse pre-trained models throughout training. This strategy aims to generate more comprehensive and representative multi- and joint-grained document representations, ultimately enhancing the effectiveness of downstream tasks related to document understanding.

\subsection{Preliminary Definitions}
Prior to going through our proposed approach in detail, we would provide formal definitions for the terminology employed throughout this paper. We believe establishing clear and precise definitions could contribute to a comprehensive understanding of the concepts and terms integral to our research.

\noindent\textbf{1) Fine-grained Document Understanding}~\cite{layoutlmv3,lilt,bros} is a pivotal aspect of document analysis, involving frameworks that offer detailed insights to comprehend document content, particularly when addressing token-level tasks, such as span-based information extraction and question answering. 
Regarding \textbf{\textit{input features}}, existing pre-trained models at the fine-grained level harness multimodal features, such as positional information and image-patch embedding, to enhance the fine-grained token representations. 
The \textbf{\textit{pre-training phase}} incorporates several learning techniques, including Masked Visual-Language Modelling, Text-Image Matching, and Multi-label Document Classification, strategically designed to acquire inter or cross-modality correlations and contextual knowledge. 
However, it is essential to acknowledge the \textbf{\textit{limitations}} of fine-grained frameworks, as their primary focus lies in learning the logical and layout arrangement of input documents. These frameworks may encounter challenges in handling complex multimodal components.

\noindent\textbf{2) Coarse-grained Document Understanding}~\cite{lxmert, visualbert} is a vital component in document analysis, with frameworks adept at grasping the logical relations and layout structures within input documents. Particularly well-suited for tasks like document component entity parsing, coarse-grained models excel in capturing high-level document understanding.
Despite the dominant trend of fine-grained document understanding models, some research recognises~\cite{lxmert, visualbert} that the knowledge from general domain-based Visual-Language Pre-trained Models (VLPMs) could be leveraged to form a foundational document understanding.
However, the coarse-grained document understanding models have significant \textbf{\textit{limitations}}, including their tendency to overlook detailed information, leading to the omission of significant words or phrases. Preliminary entity-level annotations are often necessary, and the current backbone models are pre-trained on the general domain, highlighting the need for document domain frameworks specifically pre-trained at the coarse-grained level.

\subsection[Multimodal Multi-task Multi-teacher Joint-grained Document Understanding]{Multimodal Multi-task Multi-teacher Joint-grained Document Understanding\footnote{Subsections are aligned with different colour in Figure~\ref{fig:task_definition}, Green: Section~\ref{sec:mmmt}, Blue: Section~\ref{sec:jointl}, Orange: Section~\ref{sec:mmm_loss}}}
Therefore, we introduce a joint-grained document understanding framework $\mathcal{F}_{jg}$, designed to harness pre-trained knowledge from both fine-grained and coarse-grained levels. Our approach integrates insights from multiple pre-trained backbones, facilitating a unified understanding of document content encompassing detailed nuances and high-level structures. It aims to synergise the strengths of fine-grained and coarse-grained models, enhancing the overall effectiveness of form understanding tasks. 

\subsubsection{Multimodal Multi-task Multi-Teacher}
\label{sec:mmmt}
To facilitate this joint-grained framework, we employ \textbf{M}ultimodal \textbf{M}ulti-teachers from two \textbf{M}ulti-tasks, fine-grained and coarse-grained tasks within our framework. While the fine-grained teacher $\mathcal{F}_{fg}$ is characterised by checkpoints explicitly fine-tuned for the 
\textbf{token classification}, the coarse-grained teacher $\mathcal{F}_{cg}$ utilises fine-tuning checkpoints for the 
document component \textbf{entity classification}. 
The details of fine-grained and coarse-grained teacher models are articulated in Section \ref{sec:imp_details}. The ablation study of those teacher models is in Section~\ref{sec:teacher_abl}. $\mathcal{F}_{fg}$ and $\mathcal{F}_{cg}$ get the encoded inputs of token and entity level, respectively, to acquire the corresponding last layer hidden states and logits for downstreaming procedures. For example, after feeding the sequence of tokens $\mathbf{\widetilde{t}}=\{\widetilde{t_1},\widetilde{t_2},...,\widetilde{t_k}\}$ and sequence of multimodal entity embeddings $\mathbf{\widetilde{E}}=\{\widetilde{E_1},\widetilde{E_2},...,\widetilde{E_n}\}$ into ${\mathcal{F}_{fg}}_1$ and ${\mathcal{F}_{cg}}_1$, respectively, we acquire the hidden states $\mathbf{t^1}=\{t^1_1,t^1_2,...,t^1_k\}$ and $\mathbf{E^1}=\{E^1_1,E^1_2,...,E^1_n\}$, as well as classification logits $\mathbf{p_{t^1}}=\{p_{t^1_1},p_{t^1_2},...,p_{t^1_k}\}$ and $\mathbf{p_{E^1}}=\{p_{E^1_1},p_{E^1_2},...,p_{E^1_n}\}$. Supposing $\mathbb{T} = \{\mathbf{t^1},\mathbf{t^2},...\}$  and $\mathbb{E} = \{\mathbf{E^1},\mathbf{E^2},...\}$ are hidden states from multiple teachers, the combined representations are fed into corresponding projection layers $\mathcal{L}_{fg}$ and $\mathcal{L}_{cg}$ to get the multi-teacher representations $\mathbf{\widehat{t}}=\{\widehat{t_1},\widehat{t_2},...,\widehat{t_k}\}$ and $\mathbf{\widehat{E}}=\{\widehat{E_1},\widehat{E_2},...,\widehat{E_n}\}$ for each grain. 

\subsubsection{Joint-Grained Learning}
\label{sec:jointl}
Our joint-grained learning framework comprises Joint-grained Encoder and Decoders. 

\noindent\textbf{The joint-grained encoder $\mathcal{E}$}, implemented as a transformer encoder, is designed to learn the contextual correlation between fine-grained $\mathbf{\widehat{t}}$ and coarse-grained $\mathbf{\widehat{E}}$ representations. This enables the model to capture nuanced details at the token level while simultaneously grasping the high-level structures represented by entities within the document.

\noindent\textbf{The joint-grained decoders  $\mathcal{D}$} play a crucial role in processing the augmented joint-grained representations. For the fine-grained decoder  $\mathcal{D}_{fg}$, the input comprises fine-grained token representations $\mathbf{\widehat{t}}$, with the entity representation serving as memory $\mathbf{\widehat{E}}$. This configuration allows the decoder to focus on refining and generating augmented token representations $\mathbf{t}$ based on the contextual information provided by both token and entity representations. In contrast, for coarse-grained decoder $\mathcal{D}_{cg}$, the input is the entity representation $\mathbf{\widehat{E}}$, while the memory consists of token representations $\mathbf{\widehat{t}}$. This approach enables the coarse-grained decoders to emphasise broader structures and relationships at the entity level, leveraging the memory of fine-grained token information to generate a more comprehensive entity representation $\mathbf{E}$. Overall, the proposed joint-grained architecture facilitates a comprehensive understanding of document content by incorporating fine-grained and coarse-grained perspectives.

The pre-training of different teacher models involves diverse techniques and features, so a simplistic approach of merely concatenating or pooling hidden states may not fully leverage the individual strengths of each model. Traditional self-/cross attention-based transformer encoders or decoders might encounter challenges in integrating knowledge from various grains, potentially introducing noise to specific grained weights. To address this concern, we propose using multiple types of losses to thoroughly explore implicit knowledge within the diverse teachers (pre-trained models). 

\subsection{Multimodal Multi-task Multi-Teacher Knowledge Distillation}
\label{sec:mmm_loss}
This section introduces the multi-loss strategy to enhance intra-grained and cross-grained knowledge exchange, ensuring a more nuanced and effective integration of insights from fine-grained and coarse-grained representations. The accompanying multi-loss ablation study (Section \ref{sec:loss_abl}) aims to optimise the synergies between multiple teacher models, thereby contributing to a more robust and comprehensive joint-grained learning process. 

\subsubsection{Task-oriented Cross Entropy Loss}
The Task-oriented Cross Entropy (CE) loss is pivotal in facilitating a task-based knowledge distillation strategy. This is computed by comparing the predictions of the student model with the ground truth for each specific task. Adopting the CE loss provides the student model with direct supervisory signals, thereby aiding and guiding its learning process. Note that we address two task-oriented CE losses within our proposed approach, one from the 
token classification task and the other from the 
entity classification task. The output hidden states from $\mathcal{D}_{fg}$ and $\mathcal{D}_{cg}$ are fed into classifiers to get the output logits $\mathbf{p_{t}}=\{p_{t_1},p_{t_2},...,p_{t_k}\}$ and $\mathbf{p_{E}}=\{p_{E_1},p_{E_2},...,p_{E_n}\}$. Supposing the label sets for fine-grained and entity-level tasks are $\mathbf{Y_{t}} = \{y_{t_1},y_{t_2},...,y_{t_k}\}$ and $\mathbf{Y_{E}} = \{y_{E_1},y_{E_2},...,y_{E_n}\}$, the fine-grained and coarse-grained Task-oriented Cross Entropy losses $l_t$ and $l_E$ are calculated as:
\begin{equation}
    l_{t} = Cross Entropy(\mathbf{p_{t}}, \mathbf{Y_{t}})
\end{equation}
\begin{equation}
    l_{e} = Cross Entropy(\mathbf{p_{E}}, \mathbf{Y_{E}})
\end{equation}
\subsubsection{Intra-Grained Loss Functions}
Since various pre-trained models provide different specific knowledge to understand the form comprehensively, effectively distilling valuable information from selected fine-tuned checkpoints may generate more representative token representations. In addressing this, we introduce two target-oriented loss functions tailored to distil knowledge from teachers at different levels. These aim to project the label-based distribution from fine-grained  $\mathbf{p_{\mathbb{T}}} = \{ \mathbf{p_{t^1}},\mathbf{p_{t^2}},...\}$ or coarse-grained teacher logits $\mathbf{p_{\mathbb{E}}} = \{\mathbf{p_{E^1}},\mathbf{p_{E^2}},...\}$ to corresponding student logits $\mathbf{p_{t}}$ and $\mathbf{p_{E}}$, enabling efficient learning of label distributions.

\noindent\textbf{Similarity Loss:} This is introduced as an effective method to distil knowledge from the output logits $\mathbf{p_{t}}$ and $\mathbf{p_{E}}$ of selected fine-grained or coarse-grained teacher checkpoints from $\mathbf{p_{\mathbb{T}}}$  and $\mathbf{p_{\mathbb{E}}}$. It aims to mitigate the logit differences between the student classifier and the chosen teachers using cosine similarity ($CosSim$), promoting a more aligned understanding of the label-based distribution. Supposing we have $n_t$ and $n_e$ teachers for fine-grained and coarse-grained tasks, respectively, the similarity loss of fine-grained $l_{sim_t}$ and coarse-grained $l_{sim_e}$ can calculated by:
\begin{equation}
    l_{sim_t} = - \overset{i=n_t}{\underset{i}{\Sigma}} \overset{j=k}{\underset{j}{\Sigma}} CosSim(p_{t^{i}_{j}}, p_{t_j})
\end{equation}
\begin{equation}
l_{sim_e} = - \overset{i=n_e}{\underset{i}{\Sigma}} \overset{j=n}{\underset{j}{\Sigma}} CosSim(p_{E^{i}_{j}}, p_{E_j})
\end{equation}
\noindent\textbf{Distilling Loss}: Inspired by \cite{phuong2019towards}, we adopt an extreme logit learning model for the distilling loss. This loss implements knowledge distillation using Mean Squared Error ($MSE$) between the students' logits $\mathbf{p_{t}}$ and $\mathbf{p_{E}}$  and the teachers' logit sets $\mathbf{p_{\mathbb{T}}}$  and $\mathbf{p_{\mathbb{E}}}$. This method is employed to refine the knowledge transfer process further, promoting a more accurate alignment between the student and teacher models. 
\begin{equation}
    l_{{distil}_t} = \frac{1}{k} \overset{j=k}{\underset{j}{\Sigma}} MSE(p_{t^{i}_{j}}, p_{t_j})
\end{equation}
\begin{equation}
    l_{{distil}_e} = \frac{1}{n} \overset{j=n}{\underset{j}{\Sigma}} MSE(p_{E^{i}_{j}}, p_{E_j})
\end{equation}
The introduction of these intra-grained loss functions, including the similarity loss and the distilling loss, contributes to mitigating distribution gaps and fostering a synchronised understanding of the form across various levels of granularity.

\subsubsection{Cross-Grained Loss Functions}
In addition, we incorporate cross-grained loss functions. While fine-grained and coarse-grained information inherently align, the joint-grained framework employs self-attention and cross-attention to approximate the correlation between token and entity representations. $\mathbb{T}$ and $\mathbb{E}$ are teachers hidden states sets, each $\mathbf{t^{i}} \in \mathbb{T}$ and $\mathbf{E^{i}} \in \mathbb{E}$ are represented $\mathbf{t^{i}}=\{t^i_1,t^i_2,...,t^i_k\}$ and $\mathbf{E^i}=\{E^i_1,E^i_2,...,E^i_n\}$ and  $\mathbf{t}$ and $\mathbf{E}$ are hidden states from student decoder.

\noindent\textbf{Cross-grained Triplet Loss:} Inherent in each grained feature are parent-child relations between tokens and aligned semantic form entities. The introduction of triplet loss aids the framework in automatically selecting more representative feature representations by measuring the feature distance from one grain to another-grained aligned representation. This effectively enhances joint-grained knowledge transfer, optimising the overall understanding of the form. For acquiring the loss $l_{{triplet}_{fg}}$ to select fine-grained teachers based on coarse-grained distribution adaptively, we define the anchor as each entity $E_i \in \mathbf{E}$ which has the paired token representations $t^1_i \in \mathbf{t^1}$ and $t^2_i \in \mathbf{t^2}$ (if the number of teachers is more significant than 2, randomly select two of them). The L-2 norm distance is used to measure the distance between fine-grained teachers ($t^1_i$, $t^2_i$) and anchor $E_j$, where the more similar entities are treated as positive samples ($t^{pos}_i$) otherwise negative ($t^{neg}_i$). For coarse-grained triplet loss $l_{{triplet}_{cg}}$, the same measurements are adopted for coarse-grained teacher positive ($E^{pos}_j$) and negative selection ($E^{neg}_j$) for an anchor $t_i$. Supposing the $j$-th, $l_{{triplet}_{fg}}$ and $l_{{triplet}_{cg}}$ are defined:
\begin{equation}
    l_{{triplet}_{fg}} = \frac{1}{k} \overset{i=k}{\underset{i}{\Sigma}} Triplets(E_j,t^{pos}_i,t^{neg}_i)
\end{equation}
\begin{equation}
    l_{{triplet}_{cg}} = \frac{1}{k} \overset{i=k}{\underset{i}{\Sigma}} Triplets(t_i,E^{pos}_j,E^{neg}_j)
\end{equation}
As one entity is typically paired with more than one token, when calculating $l_{{triplet}_{cg}}$, we will consider all $k$ entity-token pairs. 

\noindent\textbf{Cross-grained Alignment Loss:} In addition to the triplet loss, designed to filter out less representative teachers, we introduce another auxiliary task. This task focuses on predicting the relations between tokens and entities, providing an additional layer of refinement to the joint-grained framework. The cross-grained alignment loss further contributes to the comprehensive learning and alignment of token and entity representations, reinforcing the joint-grained understanding of the form document. For an input form document page containing $k$ tokens and $n$ entities, we have a targeting tensor $\mathbf{Y_{align}}$ where $Dim(\mathbf{{Y_{align}}}) = \mathbb{R}^{k \times n}$. We use acquired alignment logit $\mathbf{p_{align}} =\mathbf{t} \times \mathbf{E}$ to represent the predicted token-entity alignments. The cross-grained alignment loss $l_{align}$ can be calculated by:
\begin{equation}
    l_{{align}} = Cross Entropy(\mathbf{p_{align}}, \mathbf{Y_{align}})
\end{equation}

\section{Evaluation Setup}

\subsection{Datasets\footnote{The statistics of token/entity are shown in Table~\ref{tab:funsd_test} and~\ref{tab:formnlu_test}.}}
\noindent\textbf{FUNSD} \cite{funsd} comprises 199 noisy scanned documents from various domains, including marketing, advertising, and science reports related to US tobacco firms. It is split into train and test sets (149/50 documents), and each document is presented in either printed or handwritten format with low resolutions. Our evaluation focuses on the semantic-entity labeling task that identifies four predefined labels (i.e., question, answer, header, and other) based on input text content. 

\noindent\textbf{FormNLU} \cite{formnlu} consists of 867 financial form documents collected from Australian Stock Exchange (ASX) filings. It includes three form types: digital \textit{(\textbf{D})}, printed \textit{(\textbf{P})}, and handwritten \textit{(\textbf{H})}, and is split into five sets: train-\textit{\textbf{D}} (535), val-\textit{D} (76), test-\textit{D} (146), test-\textit{\textbf{P}} (50), and test-\textit{\textbf{H}} (50 documents) and supports two tasks: Layout Analysis and Key Information Extraction. Our evaluation focuses on the layout analysis that identifies seven labels (i.e., title, section, form key, form value, table key, table value, and others), detecting each document entity, especially for \textit{\textbf{P}} and \textit{\textbf{H}}, the complex multimodal form document. 

\subsection{Baselines and Metrics}
For \textbf{token-level information extraction} baselines, we use three Document Understanding (DU) models: LayoutLMv3 \cite{layoutlmv3}, LiLT \cite{lilt}, and BROS \cite{bros}. LayoutLMv3 employs a word-image patch alignment, that utilises a document image along with its corresponding text and layout position information. In contrast, LiLT and BROS focus only on text and layout information without incorporating images. LiLT uses a bi-directional attention mechanism across token embedding and layout embedding, whereas BROS uses a relative spatial encoding between text blocks. 
For \textbf{entity-level information extraction} baselines, we use two vision-language (VL) models: LXMERT \cite{lxmert} and VisualBERT \cite{visualbert}. Compared to the two DU models, these VL models use both image and text input without layout information. LXMERT focuses on cross-modality learning between word-level sentence embeddings and object-level image embeddings, while VisualBERT simply inputs image regions and text, relying on implicit alignments within the network. 
For \textbf{evaluation metrics}, inspired by \cite{funsd} and \cite{formnlu}, we primarily use F1-score to represent both overall and detailed performance breakdowns, aligning with other baselines. 

\subsection{Implementation Details\footnote{Additional Implemtnation Details are in Appendix~\ref{sec:num_para}}}
\label{sec:imp_details}
In token-level experiments, we fine-tuned LayoutLMv3-base using its text tokeniser and image feature extractor. We also fine-tuned LiLT combined with RoBERTa base. In entity-level experiments, we employ pre-trained BERT (748-d) for encoding textual content, while ResNet101(2048-d) is used for region-of-interest(RoI) feature to capture the visual aspect. These extracted features serve as input for fine-tuning LXMERT and VisualBERT. All fine-tuned models serve as teacher models. Our hyperparameter testing involves a maximum of 50 epochs with learning rates set at 1e-5 and 2e-5. All are conducted on a Tesla V100-SXM2 with 16GB graphic memory and 51 GB memory, CUDA 11.2.

\section{Results}
\subsection{Overall Performance}
\label{sec:overall_perform}
\begin{table}[t]
\centering
\begin{adjustbox}{max width =0.95\linewidth}

\begin{tabular}{l|l|c|cc}
\hline

\multirow{2}{*}{\textbf{Model}} & \multirow{2}{*}{\textbf{Config \& Loss}} & \multirow{2}{*}{\textbf{FUNSD}} & \multicolumn{2}{c}{\textbf{FormNLU}} \\ 
\cline{4-5}
& & &\textbf{\textit{P}}&\textit{\textbf{H}}\\\hline
\hline

BROS & Single Teacher & 82.44 & 92.45 & 93.68 \\
LiLT & Single Teacher & 87.54 & \underline{96.50} & 91.35 \\
LayoutLMv3 & Single Teacher & \underline{90.61} & 95.99 & \underline{97.39} \\
\hline
\hline
JG-$\mathcal{E}$ & Joint Cross Entropy  & 90.45 & 94.91 & 96.55 \\
JG-$\mathcal{D}$  & Joint Cross Entropy  & 90.48 & 95.68 & 97.62 \\
JG-$\mathcal{E}$\&$\mathcal{D}$ & Joint Cross Entropy  & 90.57 & 95.93 & 97.62 \\
\hline
\hline
\multirow{7}{*}{MT-JG-$\mathcal{E}$\&$\mathcal{D}$} & Joint Cross Entropy & 90.53 & 97.21 &97.75 \\
\cline{2-5}
  & + \textit{Sim}  & \textbf{91.05} & 98.25 & {98.09} \\
 &  + \textit{Distil}  & 90.90 & 98.12 & 97.72 \\
\cline{2-5}
 &  + \textit{Triplet} & 90.28 & 97.58 & 97.28 \\
 (Ours) & + \textit{Align} & 90.55 & 97.24 & 97.42 \\
 \cline{2-5}
 & + \textit{Sim} + \textit{Distil} & \multirow{2}{*}{{90.92}} & \multirow{2}{*}{{\textbf{98.69}}} & \multirow{2}{*}{{\textbf{98.39}}}  \\ 
 & + \textit{Triplet} + \textit{Align} & & & \\
 \hline
\end{tabular}
\end{adjustbox}
\caption{Overall performance with configurations on FormNLU printed \textbf{\textit{P}} and handwritten \textbf{\textit{H}}. 
The full form of acronyms can be found in Section \ref{sec:overall_perform}. The best is in \textbf{bold}.
The best teacher model (baseline) is \underline{underlined}.}
\label{tab:overall_performance}

\end{table}

Extensive experiments are conducted to highlight the effectiveness of the proposed \textbf{Multimodal Multi-task Multi-Teacher} framework, including \textit{joint-grained learning}, \textit{multi-teacher} and \textit{multi-loss} architecture. Table~\ref{tab:overall_performance} shows representative model configurations on various adopted modules. LayoutLMv3 performs notably superior to BROS and LiLT, except for the FormNLU printed test set. LayoutLMv3 outperforms around 3\% and 4\% the second-best baseline on FUNSD and FormNLU handwritten sets, respectively. This superiority can be attributed to LayoutLMv3's utilisation of patched visual cues and textual and layout features, resulting in more comprehensive multimodal representations. So we found LayoutLMv3 would be a robust choice for fine-grained baselines in further testing\footnote{We chose LLmv3 and LXMERT for JG and
select LLMv3\&LilT and VBERT\&LXMERT for MT-JG-$\mathcal{E}\&\mathcal{D}$. More teacher combinations analysis is in Section~\ref{sec:teacher_abl}.}. 
To find the most suitable \textbf{Joint-Grained learning} (JG), we compare the results of single-teacher joint-grained frameworks including Encoder ($\mathcal{E}$) only, Decoder ($\mathcal{D}$) only, and Encoder with Decoder ($\mathcal{E}\&\mathcal{D}$). Table~\ref{tab:overall_performance} illustrates $\mathcal{E}\&\mathcal{D}$ achieving the highest performance among three baselines. However, upon integrating multiple teachers from each grain (MT-JG-$\mathcal{E}\&\mathcal{D}$), competitive performance is observed compared to the baselines on both FormNLU printed (\textbf{\textit{P}}) (from LiLT 96.5\% to 97.21\%) and handwritten set (\textbf{\textit{H}}) (from LiLT 97.39\% to 97.75\%).  
Still, additional techniques may be necessary to distil the cross-grained multi-teacher information better.

To thoroughly distil joint-grained knowledge from multiple teachers, we introduced multiple loss functions encompassing \textbf{Multiple auxiliary tasks}. These functions capture teacher knowledge from intra-grained and cross-grained perspectives, generating representative token embeddings.  
Typically, using either intra-grained or coarse-grained loss individually leads to better performance than the best baselines across various test sets. Intra-grained Similarity (\textit{Sim}) and Distilling (\textit{Distil}) loss consistently achieve higher F1 scores in nearly all test sets. Moreover, cross-grained \textit{Triplet} and alignment (\textit{Align}) losses outperform the best baseline on the FormNLU (\textbf{\textit{P}}) or (\textbf{\textit{H}}). This highlights the effectiveness of the proposed multi-task learning approach in enhancing token representations by integrating knowledge from joint-grained multi-teachers. 
Intra-grained loss functions exhibit higher robustness on both datasets, whereas cross-grained loss functions only perform well on FormNLU. This difference may stem from the FUNSD being sourced from multiple origins, whereas FormNLU is a single-source dataset. Coarse-grained loss functions may excel on single-source documents by capturing more prevalent knowledge but might introduce noise when applied to multiple sources. Also, the model demonstrates its most competitive performance by integrating all proposed loss functions (\textit{+Sim+Distil+Triplet+Align}). This highlights how the proposed intra-grained and cross-grained loss functions enhance multi-teacher knowledge distillation in form understanding tasks\footnote{More loss combination analysis is in Section~\ref{sec:loss_abl}}. 

\begin{figure*}
\centering
\includegraphics[width=\linewidth]{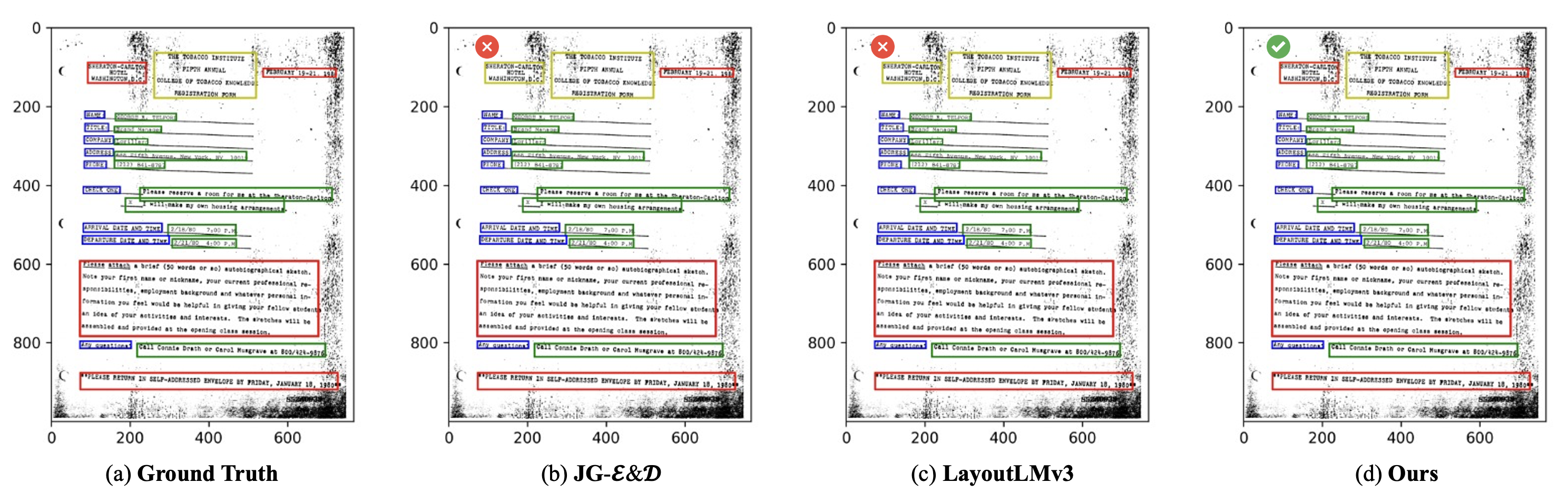}
\caption{Example output showing (a) Ground Truth (b) JG-$\mathcal{E}$\&$\mathcal{D}$ (c) LayoutLMv3, and (d) Ours on a FUNSD page. The color code for layout component labels is as follows; \textcolor{blue}{Question}, \textcolor{teal}{Answer}, \textcolor{olive}{Header}, \textcolor{red}{Other}. Our model, employing the best loss combination (cross-entropy + similarity) on FUNSD, accurately classified all layout components.}
\label{fig:FUNSD_case}
\end{figure*}


\subsection{Effect of Multi-Teachers}
\label{sec:teacher_abl}
\begin{table}[t]
\centering
\begin{adjustbox}{max width =0.95\linewidth}

\begin{tabular}{c|c||c|c|c}
\hline
\multirow{2}{*}{\textbf{FG Teacher}} & \multirow{2}{*}{\textbf{CG Teacher}} & \multirow{2}{*}{\textbf{FUNSD}} & \multicolumn{2}{c}{\textbf{FormNLU}} \\ \cline{4-5} 
 & & & \textit{\textbf{P}} & \textbf{H}\\ \hline \hline
\multirow{3}{*}{LLmv3} & VBERT & 90.19 & 94.72 & 96.99 \\ \cline{2-5} 
 & LXMERT & \textbf{90.57} & 95.93 & \underline{97.62} \\ \cline{2-5} 
 & Transformer & 90.22 & 93.65 & 95.94 \\ \hline
\multirow{3}{*}{LiLT} & VBERT & 87.66 & \textbf{97.65} & 90.53 \\ \cline{2-5} 
 & LXMERT & 87.34 & 96.76 & 91.18 \\ \cline{2-5} 
 & Transformer & 87.91 & 97.20 & 90.58 \\ \hline
LLmv3 & VBERT\&LXMERT & 90.42 & 95.05 & 97.25 \\ \hline
LLmv3 \& LiLT&LXMERT & 90.39 & 96.73 & 97.42 \\ \hline

\textbf{LLmv3\&LiLT} & \textbf{VBERT\&LXMERT} & \underline{90.53} & \underline{97.21} &\textbf{ 97.75} \\ \hline
\end{tabular}
\end{adjustbox}
\caption{Comparison of Performance across Teacher Combinations. FG: Fine-Grained, CG: Coarse-Grained, LLmv3: LayoutLMv3, VBERT: VisualBERT. The best is in \textbf{bold}. The second best is \underline{underlined}. This ablation study is based on only Joint Cross Entropy Loss.}
\label{tab:teacher_ablation}
\end{table}

We analysed various teacher combinations to ensure they provide sufficient knowledge for improving joint-grained representations, as depicted in Table~\ref{tab:teacher_ablation}. For fine-grained teachers, since BROS underperforms compared to others, we only include the performance of its counterparts. The LayoutLMv3-based joint framework performs better, outperforming LiLT-based by approximately 3\% on FUNSD and over 5\%  on FormNLU (\textbf{\textit{H}}). This improvement can be attributed to the contextual learning facilitated by visual cues. Notably, LiLT achieves the highest performance on the FormNLU (\textbf{\textit{P}}), likely due to its well-designed positional-aware pre-training tasks.
For coarse-grained teachers, pre-trained backbones demonstrate better robustness than randomly initialised Transformers, highlighting the benefits of general domain pre-trained knowledge in form understanding tasks.
Table~\ref{tab:teacher_ablation} illustrates multiple teachers cannot always ensure the best performance, however, the robustness of the proposed model is enhanced by capturing more implicit knowledge from cross-grained teachers.
\subsection{Effect of Loss Functions}
\label{sec:loss_abl}

\begin{table}[t]
\centering
\begin{adjustbox}{max width =1\linewidth}

\begin{tabular}{|cccc||c|cc}
\hline
\multicolumn{4}{c||}{\textbf{Loss Functions}}                                                                                       & \multirow{2}{*}{\textbf{FUNSD}}    & \multicolumn{2}{c}{\textbf{FormNLU}}                                                                                                                  \\ \cline{1-4} \cline{6-7} 
\multicolumn{1}{c|}{\textbf{Similarity}} & \multicolumn{1}{c|}{\textbf{Distiling}} & \multicolumn{1}{c|}{\textbf{Triplet}} & \textbf{Alignment} &                                    & \multicolumn{1}{c|}{\textbf{\textit{\textbf{P}}}}  & {\textbf{\textit{\textbf{H}}}} \\ \hline \hline
\multicolumn{1}{c|}{O}            & \multicolumn{1}{c|}{X}               & \multicolumn{1}{c|}{X}                & X              & \textbf{91.05}    & \multicolumn{1}{c|}{98.25}                                                           & 98.09                                                           \\ \hline
\multicolumn{1}{c|}{X}            & \multicolumn{1}{c|}{O}               & \multicolumn{1}{c|}{X}                & X              & 90.90                              & \multicolumn{1}{c|}{98.12}                                                           & 97.72                                                           \\ \hline
\multicolumn{1}{c|}{X}            & \multicolumn{1}{c|}{X}               & \multicolumn{1}{c|}{O}                & X              & 90.28                              & \multicolumn{1}{c|}{97.58}                                                           & 97.28                                                           \\ \hline
\multicolumn{1}{c|}{X}            & \multicolumn{1}{c|}{X}               & \multicolumn{1}{c|}{X}                & O              & 90.55                              & \multicolumn{1}{c|}{97.24}                                                           & 97.42                                                           \\ \hline
\multicolumn{1}{c|}{O}            & \multicolumn{1}{c|}{O}               & \multicolumn{1}{c|}{X}                & X              & 90.63                              & \multicolumn{1}{c|}{98.53}                                                           & 97.22                                                           \\ \hline
\multicolumn{1}{c|}{O}            & \multicolumn{1}{c|}{X}               & \multicolumn{1}{c|}{O}                & X              & 90.51                              & \multicolumn{1}{c|}{97.71}                                                           & 97.79                                                           \\ \hline
\multicolumn{1}{c|}{O}            & \multicolumn{1}{c|}{X}               & \multicolumn{1}{c|}{X}                & O              & 90.82                              & \multicolumn{1}{c|}{97.80}                                                           & 98.05                                                           \\ \hline
\multicolumn{1}{c|}{X}            & \multicolumn{1}{c|}{O}               & \multicolumn{1}{c|}{O}                & X              & 90.82                              & \multicolumn{1}{c|}{98.22}                                                           & \underline{98.35}                              \\ \hline
\multicolumn{1}{c|}{X}            & \multicolumn{1}{c|}{O}               & \multicolumn{1}{c|}{X}                & O              & 90.83                              & \multicolumn{1}{c|}{98.63}                                                           & 97.45                                                           \\ \hline
\multicolumn{1}{c|}{O}            & \multicolumn{1}{c|}{O}               & \multicolumn{1}{c|}{O}                & X              & 90.79                              & \multicolumn{1}{c|}{98.56}                                                           & 97.72                                                           \\ \hline
\multicolumn{1}{c|}{O}            & \multicolumn{1}{c|}{O}               & \multicolumn{1}{c|}{X}                & O              & 90.66                              & \multicolumn{1}{c|}{\textbf{98.72}}                                 & 97.85                                                           \\ \hline
\multicolumn{1}{c|}{O}            & \multicolumn{1}{c|}{O}               & \multicolumn{1}{c|}{O}                & O              & \underline{90.92} & \multicolumn{1}{c|}{\underline{98.69}}                              & \textbf{98.39}                                 \\ \hline
\end{tabular}
\end{adjustbox}

\caption{Performance comparison across loss functions. The best is in \textbf{bold}. The second best is \underline{underlined}.}
\label{tab:loss_abl}
\vspace{-1em}
\end{table}

To comprehensively investigate the impact of different loss functions and their combinations, we present the performance of various combinations in Table~\ref{tab:loss_abl}. While employing intra-grained loss individually often proves more effective than using cross-grained loss alone, combining the two losses can enhance knowledge distillation from joint-grained multi-teachers. For instance, concurrently employing distilling(Distil) and Triplet loss improved accuracy from 97.72\% to 98.35\%.  Notably, stacking all proposed loss functions resulted in the best or second-best performance across all test sets, showcasing their effectiveness in distilling knowledge from multi-teacher to student models for generating more representative representations. Even though cross-grained Triplet and Alignment losses were ineffective individually, when combined with intra-grained loss, they significantly improved knowledge distillation effectiveness.

\subsection{Qualitative Analysis: Case Studies\footnote{A Case Study for FormNLU can be found in Figure~\ref{fig:FormNLU_case}}}
We visualised the sample results for the top 3 - Our best model with the best configuration, the best baseline~LayoutLMv3 and the second best baseline~\textit{JG-$\mathcal{E}$\&$\mathcal{D}$} of FUNSD in Figure~\ref{fig:FUNSD_case}. We can see that both ~LayoutLMv3 and ~\textit{JG-$\mathcal{E}$\&$\mathcal{D}$} have wrongly recognised an \textit{Other} (marked by a white cross in red circle), whereas ours has accurately recognised all document tokens and components. 

\section{Conclusion}
We introduced a Multimodal Multi-task Multi-Teacher framework in Visually-Rich form documents. Our model incorporates \textit{multi-teacher}, \textit{multi-task}, and \textit{multi-loss}, and the results show the robustness in capturing implicit knowledge from multi-teachers for understanding diverse form document natures, such as scanned, printed, and handwritten. We hope our work provides valuable insights into leveraging multi-teacher and multi-loss strategies for document understanding research.

\section*{Limitations}
\textbf{Benchmark Scope}:
Despite the paramount importance of document understanding across various domains such as finance, medicine, and resources, our study is constrained by the limited availability of visually-rich form document understanding datasets, particularly those of high quality. In this research, we solely rely on publicly available English-based form document understanding datasets. The scope of benchmark datasets, therefore, may not comprehensively represent the diversity and complexity present in form documents across different languages and industries.

\noindent\textbf{Availability of Document Understanding Teachers}:
The current limitation stems from the reliance on general document understanding teacher models due to the absence of large pre-trained form-specific document models. The availability of high-quality teachers specifically tailored for form document understanding is crucial. Future advancements in the field would benefit from the development of dedicated pre-trained models for form document understanding, providing more accurate knowledge transfer during training.

\bibliography{anthology}
\bibliographystyle{acl_natbib}

\clearpage
\appendix

\section{Statistics of tokens and entities}
\label{sec:dataset_statistics}
The following Table~\ref{tab:funsd_test} and \ref{tab:formnlu_test} demonstrates the number of tokens(length) and number of document entities. While FUNSD has 4 types(Question, Answer, Header, Other) of document entities, FormNLU has 7 types(Title, Section, Form Key, Form Value, Table Key, Table Value, Other). For the FormNLU, we applied two types of test set, including Printed~\textbf{P} and Handwritten~\textbf{H}.

\begin{figure*}[t]
    \centering
    \includegraphics[width=\linewidth]{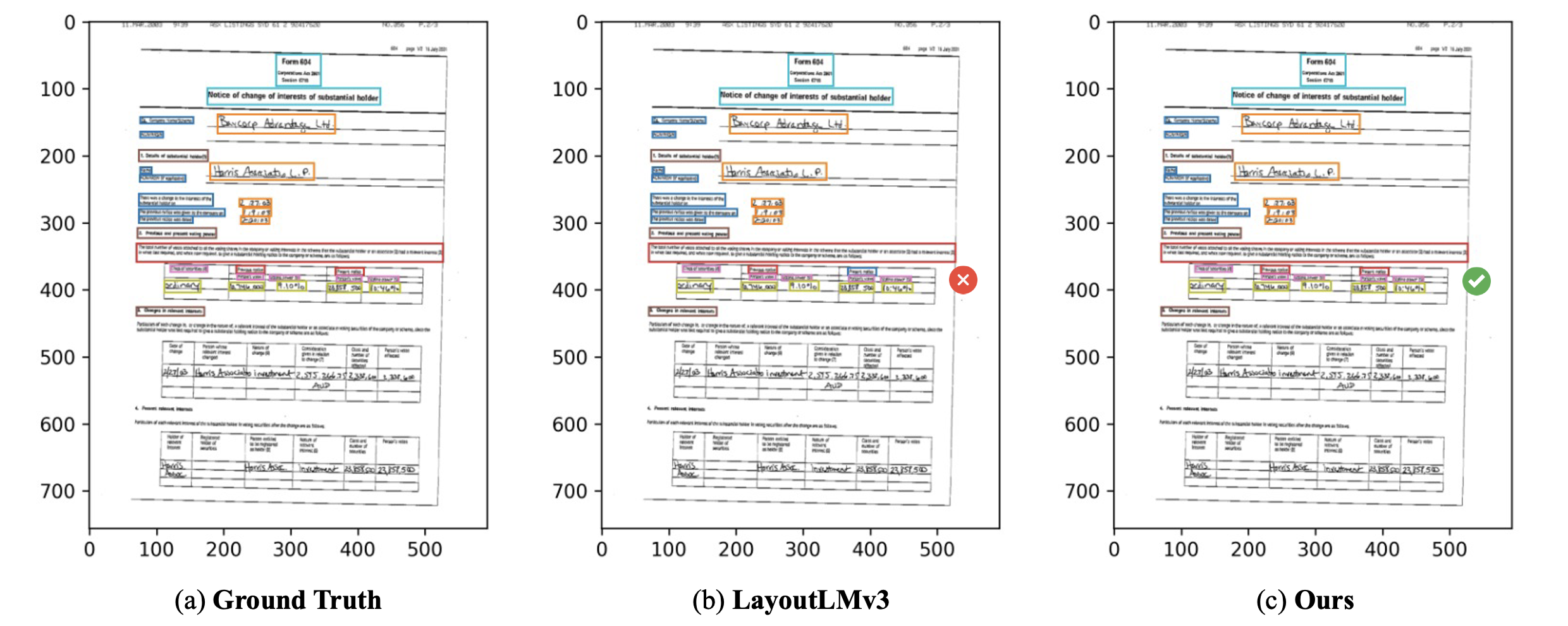}
    \caption{Example output showing (a) Ground Truth (b) LayoutLMv3, and (c) Ours on a FormNLU handwritten test set. The color code for layout component labels is as follows; \textcolor{cyan}{Title}, \textcolor{brown}{Section}, \textcolor{blue}{Form Key}, \textcolor{orange}{Form Value}, \textcolor{magenta}{Table Key}, \textcolor{olive}{Table Value}, \textcolor{red}{Other}. Our model, the best loss combination (+Sim+Distil+Triplet+Align) on FormNLU~\textbf{H}, accurately classified all layout components.}
    \label{fig:FormNLU_case}
\end{figure*}


\begin{table}[ht]
\centering
\begin{adjustbox}{max width =0.8\linewidth}
\begin{tabular}{l|l|l|l|l|l}
\hline
\textbf{\begin{tabular}[c]{@{}l@{}}FUNSD \\ (Testing)\end{tabular}} & \textbf{Question} & \textbf{Answer} & \textbf{Header} & \textbf{Other} & \textbf{Total} \\ \hline
\textbf{Entity} & 1077 & 821 & 122 & 312 & \textbf{2332} \\ \hline
\textbf{Token} & 2654 & 3294 & 374 & 2385 & \textbf{8707} \\ \hline
\end{tabular}
\end{adjustbox}
\caption{FUNSD Testing Dataset Distribution by Label.}
\label{tab:funsd_test}
\end{table}

\begin{table}[h]
\centering
\begin{adjustbox}{max width =\linewidth}
\begin{tabular}{ll|r|r|r|r|r|r|r|r}
\hline
\multicolumn{2}{l|}{\textbf{\begin{tabular}[c]{@{}l@{}}FormNLU \\ (Testing)\end{tabular}}} & \multicolumn{1}{l|}{\textbf{Title}} & \multicolumn{1}{l|}{\textbf{Section}} & \multicolumn{1}{l|}{\begin{tabular}[c]{@{}l@{}}\textbf{Form} \\ \textbf{Key}\end{tabular}} & \multicolumn{1}{l|}{\begin{tabular}[c]{@{}l@{}}\textbf{Form} \\ \textbf{Value}\end{tabular}} & \multicolumn{1}{l|}{\begin{tabular}[c]{@{}l@{}}\textbf{Table}\\ \textbf{Key}\end{tabular}} & \multicolumn{1}{l|}{\begin{tabular}[c]{@{}l@{}}\textbf{Table} \\ \textbf{Value}\end{tabular}} & \multicolumn{1}{l|}{\textbf{Others}} & \multicolumn{1}{l}{\textbf{Total}} \\ \hline
\multicolumn{1}{l|}{\textit{\textbf{P}}} & \multirow{2}{*}{\textbf{Entity}} & 98 & 100 & 346 & 332 & 250 & 249 & 152 & \textbf{1527} \\ \cline{1-1} \cline{3-10} 
\multicolumn{1}{l|}{\textit{\textbf{H}}} &  & 100 & 100 & 348 & 315 & 249 & 226 & 149 & \textbf{1487} \\ \hline
\multicolumn{1}{l|}{\textit{\textbf{P}}} & \multirow{2}{*}{\textbf{Token}} & 700 & 1258 & 1934 & 1557 & 993 & 389 & 3321 & \textbf{10152} \\ \cline{1-1} \cline{3-10} 
\multicolumn{1}{l|}{\textit{\textbf{H}}} &  & 742 & 1031 & 1805 & 866 & 779 & 366 & 2918 & \textbf{8507} \\ 
\hline
\end{tabular}
\end{adjustbox}
\caption{FormNLU Testing Dataset Distribution by Label, where \textbf{\textit{P}} and \textbf{\textit{H}} are printed and handwritten sets.}
\label{tab:formnlu_test}
\end{table}

%
\section{Breakdown Result Analysis}

\begin{table}[ht]
\centering
\begin{adjustbox}{max width =0.95\linewidth}
\begin{tabular}{l|l|l|c|c|c}
\hline
\multirow{2}{*}{\textbf{Model}} & \multirow{2}{*}{\textbf{Config}} & \multirow{2}{*}{\textbf{Overall}} & \multicolumn{3}{c}{\textbf{Breakdown}} \\ 
\cline{4-6}
 &  &  & \textbf{Header} & \textbf{Question} & \textbf{Answer} \\ \hline
LiLT & Teacher & 87.54 & 55.61 & 90.20 & 88.34 \\ \hline
LayoutLMv3 & Teacher & 90.61 & \underline{66.09} & 91.60 & \textbf{92.78} \\ \hline
JG-$\mathcal{E}$ & Joint CE & 90.45 & 64.94 & 91.70 & 92.67 \\ \hline
JG-$\mathcal{D}$ & Joint CE & 90.48 & 64.07 & 91.58 & \underline{92.73} \\ \hline
JG-$\mathcal{E}$\&$\mathcal{D}$ & Joint CE & 90.57 & 64.66 & 91.48 & \underline{92.73} \\ \hline
\multirow{7}{*}{MT-JG-$\mathcal{E}$\&$\mathcal{D}$} & Joint CE & 90.53 & 61.24 & 92.40 & 91.75 \\ 
 \cline{2-6}
 & Sim & \textbf{91.05} & 64.81 & \underline{92.58} & 92.46 \\ 
 & Distil & 90.90 & \textbf{66.96} & \textbf{92.61} & 91.97 \\ 
  \cline{2-6}
 & Triplet & 90.28 & 62.44 & 92.00 & 91.44 \\ 
 & Align & 90.55 & 63.81 & 91.82 & 92.29 \\ 
 \cline{2-6}

  & +Sim+Distil & \multirow{2}{*}{\underline{90.92}} & \multirow{2}{*}{64.22} & \multirow{2}{*}{92.54} & \multirow{2}{*}{92.31} \\ 
 & +Triplet+Align & & & \\
 \hline
\end{tabular}
\end{adjustbox}
\caption{Breakdown Results of FUNSD dataset.}
\label{tab:funsd_breakdown}
\end{table}
As shown in Table~\ref{tab:funsd_breakdown}, for the FUNSD dataset, we could find all Joint-Grained(JG-) frameworks can have a delicate performance on recognising \textit{Question} and \textit{Answer}, but decreased in Header classification. This might result from the limited number of \textit{Headers} in the FUNSD, leading to inadequate learning of the fine-grained and coarse-grained \textit{Header} information. Multi-task-oriented intra-grained and coarse-grained functions can increase the performance of \textit{Question} recognition by boosting the knowledge distilling from joint-grained multi-teachers. Especially, intra-grained knowledge distillation methods can achieve around 1\% higher than LayoutLMv3. The FUNSD dataset cannot illustrate the benefits of cross-grained loss functions well. 
\begin{table*}[h]
\centering
\begin{adjustbox}{max width =\linewidth}
\begin{tabular}{l|l|c|c|c|c|c|c|c|c|c|c|c|c|c|c}
\hline
\multirow{2}{*}{\textbf{Model}} & \multirow{2}{*}{\textbf{Config}} & \multicolumn{7}{c}{\textbf{FormNLU Printed Overall and Breakdown}} &\multicolumn{7}{|c}{\textbf{FormNLU Handwritten Overall and Breakdown}} \\
\cline{3-16}
 &  & \textbf{ Overall} & \textbf{ Sec} & \textbf{ Title} & \textbf{ F\_K} & \textbf{ F\_V} & \textbf{ T\_K} & \textbf{ T\_V} & \textbf{ Overall} & \textbf{ Sec} & \textbf{ Title} & \textbf{ F\_K} & \textbf{ F\_V} & \textbf{ T\_K} & \textbf{ T\_V} \\ \hline
LiLT & Teacher & 96.50 & 98.32 & 96.97 & 98.84 & 96.62 & 96.57 & 93.60 & 91.35 & 95.39 & 99.50 & 94.81 & 90.67 & 84.19 & 89.81 \\ \hline
LayoutLMv3 & Teacher & 95.99 & 98.45 & 97.96 & 97.97 & 96.73 & 92.37 & 92.98 & 97.39 & \textbf{99.33} & 99.01 & \textbf{99.85} & 98.24 & 93.95 & 95.95 \\ \hline
JG-$\mathcal{E}$ & Joint CE & 94.91 & \textbf{99.66} & 98.99 & 98.11 & 95.73 & 90.14 & 90.31 & 96.55 & \textbf{99.33} & 99.01 & 99.42 & \textbf{98.56} & 88.37 & 94.67 \\ \hline
JG-$\mathcal{D}$ & Joint CE & 95.68 & \textbf{99.66} & \textbf{100.00} & 98.55 & 96.45 & 91.94 & 91.10 & 97.62 & \textbf{99.33} & 99.01 & \textbf{99.85} & \textbf{98.56} & 93.02 & 95.98 \\ \hline
JG-$\mathcal{E}$\&$\mathcal{D}$ & Joint CE & 95.93 & \textbf{99.66} & 97.96 & 97.82 & 97.18 & 91.97 & 92.15 & 97.62 & \textbf{99.33} & 99.01 & \textbf{99.85} & 98.40 & 93.74 & 95.75 \\ \hline
\multirow{7}{*}{MT-JG-$\mathcal{E}$\&$\mathcal{D}$} & Joint CE & 97.21 & 99.32 & 98.48 & 99.57 & 96.58 & 97.35 & 95.06 & 97.75 & 97.67 & 99.50 & 99.13 & 97.93 & 95.55 & 96.41 \\ 
\cline{2-16}
 & Sim & \underline{98.25} & 99.32 & 99.49 & 99.28 & 97.75 & \textbf{97.96} & \textbf{97.12} & \underline{98.09} & 99.00 & \textbf{100.00} & 99.27 & 98.25 & \underline{96.45} & 96.61 \\ 
 & Distil & 98.12 & 99.32 & \textbf{100.00} & \textbf{99.71} & \underline{97.90} & \underline{97.55} & 96.30 & 97.72 & 97.35 & \textbf{100.00} & 99.13 & 97.62 & 95.75 & 97.07 \\ 
 \cline{2-16}
 & Triplet & 97.58 & 99.32 & 99.49 & 99.28 & 97.18 & \underline{97.55} & 95.87 & 97.28 & 98.00 & \textbf{100.00} & 98.83 & 97.31 & 93.90 & 96.83 \\ 
 & Align & 97.24 & 99.32 & 98.48 & \textbf{99.71} & 96.57 & 96.13 & 95.47 & 97.42 & \textbf{99.33} & 99.50 & 99.13 & 96.85 & 92.86 & \underline{97.52 }\\ 
\cline{2-16}
 & +Sim+Distil & \multirow{2}{*}{\textbf{98.69}} & \multirow{2}{*}{99.32} & \multirow{2}{*}{\textbf{100.00}} & \multirow{2}{*}{\textbf{99.71}} & \multirow{2}{*}{\textbf{99.25}} & \multirow{2}{*}{97.35} & \multirow{2}{*}{\textbf{97.12}} & \multirow{2}{*}{\textbf{98.39}} & \multirow{2}{*}{98.33} & \multirow{2}{*}{\textbf{100.00}} & \multirow{2}{*}{99.56} & \multirow{2}{*}{98.09} & \multirow{2}{*}{\textbf{96.94}} & \multirow{2}{*}{\textbf{97.75}} \\ 
 & +Triplet+Align & & & & & &&&&&&&& \\\hline
\end{tabular}
\end{adjustbox}
\caption{Overall and Breakdown Analysis of FormNLU Printed Set and Handwritten Set. The categories of FormNLU dataset Task A include Section (Sec), Title, Form\_Key (F\_K), Form\_Value (F\_V), Table\_Key (T\_K), Table\_Value (T\_V).}
\label{tab:comparison}
\end{table*}

For FormNLU printed and handwritten sets, the joint-grained framework and proposed loss functions can effectively improve \textit{Section} (\textit{Sec}) and Title recognition. As the \textit{Title}, \textit{Section} and \textit{Form\_key} (\textit{F\_K}) are normally located at similar positions for single-source forms, this may demonstrate both joint-grained framework and multi-task loss function could distil knowledge. Additionally, baseline models are not good at recognising table keys and values, especially handwritten sets. As we use the layoutLMv3 in the joint-grained framework, the performance of recognising table-related tokens is not good for the joint-learning framework. After integrating multiple teachers, the performance has increased from 91.97\% to 97.35\%  on the printed set. The proposed multi-task loss functions may achieve a higher performance of 97.96\%. Significant improvements can also be observed across two test sets across all table-related targets. This illustrates that the joint-grained multi-teacher framework can effectively tackle the limitation of one teacher to generate more comprehensive token representations, and the intra-grained and cross-grained loss could boost the effective knowledge exchange to make the generalisation and robustness of the entire framework. 

\section{Additional Qualitative Analysis}
In our qualitative evaluation, we took a closer look at the results by visualising the output of the top two models—our best-performing model with the optimal configuration and the baseline \textit{LayoutLM3}—on the FormNLU handwritten set, as presented in Figure~\ref{fig:FormNLU_case}. This examination revealed a notable discrepancy between the models. Specifically, \textit{LayoutLM3} exhibited an erroneous identification of the Table Key as a Form Key. In contrast, our model demonstrated a higher level of precision by accurately recognising and distinguishing all components within this intricate and noise-laden handwritten document.

This illustrative case serves as a compelling example highlighting the challenges associated with relying solely on knowledge from a single document to understand teachers. The complexity of distinguishing various document structures, such as the nuanced difference between a form key and a table key, becomes evident. The inadequacy of a singular teacher's knowledge in capturing such intricacies emphasises the importance of our proposed \textbf{Multi-modal Multi-task Multi-Teacher} framework, which leverages insights from multiple teachers to enhance the robustness and accuracy of form document understanding.

\section{Additional Implementation Details}
\label{sec:num_para}
The table presented in Table~\ref{tab:num_parameters} outlines the number of total parameters and trainable parameters across various model configurations. It is evident that the choice of teacher models primarily determines the total number of parameters. As the number of teachers increases, there is a corresponding enhancement in the total parameter count.
Furthermore, the architecture of the student model significantly influences the number of trainable parameters. For instance, encoder-decoder-based student models exhibit a higher count of trainable parameters compared to architectures employing only an encoder or decoder. This discrepancy implies that training encoder-decoder models demands more computational resources.
Despite the variation in trainable parameters among different student model architectures, it is noteworthy that the overall number remains substantially smaller than that of single-teacher fine-tuning processes. This observation underscores the efficiency of student model training in comparison to fine-tuning pre-trained models.

\begin{table}[t]
\centering
\begin{adjustbox}{max width =\linewidth}

\begin{tabular}{l|c|c|c|l}
\hline
\textbf{Fine-grained} & \textbf{Coarse-Grained} & \textbf{Configure} & \textbf{\# Para} & \textbf{\# Trainable} \\
\hline
LiLT & N/A & Teacher & 130,169,799 & 130,169,799 \\ \hline
\multirow{5}{*}{LayoutLMv3}  & N/A & Teacher & 125,332,359 & 125,332,359 \\ \cline{2-5}
& \multirow{3}{*}{LXMERT} & JG-Encoder & 393,227,514 & 19,586,415 \\
\cline{3-5}
 &  & JG-Decoder & 423,952,890 & 50,311,791 \\
 \cline{3-5}
 &  & \multirow{4}{*}{JG-$\mathcal{E}$\&$\mathcal{D}$}  & 440,494,842 & 66,853,743 \\
 \cline{2-2}
 \cline{4-5}
 & VisualBERT\&LXMERT & & 557,260,798 & 70,394,991 \\
\cline{1-2}
 \cline{4-5}
\multirow{2}{*}{LayoutLMv3\&LiLT} & LXMERT & & 574,205,889 & 68,034,159 \\
& VisualBERT\&LXMERT & & 688,611,013 & 71,575,407 \\
\hline
\end{tabular}
\end{adjustbox}
\caption{Model configurations and parameters}
\label{tab:num_parameters}
\end{table}
\end{document}